\documentclass{article}
\usepackage[utf8]{inputenc}
\usepackage{graphicx}
\usepackage{algorithm}
\usepackage{algpseudocode}
\usepackage{comment}
\usepackage{enumitem}
\usepackage{makecell}
\usepackage{siunitx}
\usepackage{colortbl}
\usepackage[numbers]{natbib}
\usepackage{array}
\usepackage{amsmath,amssymb,amsfonts}

\usepackage{textcomp}
\usepackage{subfigure}

\usepackage{fancyhdr}
\title{Unmanned Aerial Vehicle Control Through Domain-based Automatic Speech Recognition}

\author{
Ruben Contreras$^{1}$ \and 
Angel Ayala$^{2}$ \and 
Francisco Cruz$^{1,3}$} 

\date{\normalsize
$^{1}$ Escuela de Ingenier\'ia, Universidad Central de Chile, Santiago, Chile.\\
$^{2}$ Escola Polit\'ecnica de Pernambuco, Universidade de Pernambuco,\\Recife, Brasil.\\
$^{3}$ School of Information Technology, Deakin University, Geelong, Australia.\\
Corresponding e-mails: ruben.contreras@alumnos.ucentral.cl, aaam@ecomp.poli.br, francisco.cruz@deakin.edu.au\\
}

\begin{document}

\maketitle

\begin{abstract}
Currently, unmanned aerial vehicles, such as drones, are becoming a part of our lives and reaching out to many areas of society, including the industrialized world. 
A common alternative to control the movements and actions of the drone is through unwired tactile interfaces, for which different remote control devices can be found. 
However, control through such devices is not a natural, human-like communication interface, which sometimes is difficult to master for some users.
In this work, we present a domain-based speech recognition architecture to effectively control an unmanned aerial vehicle such as a drone.
The drone control is performed using a more natural, human-like way to communicate the instructions.
Moreover, we implement an algorithm for command interpretation using both Spanish and English languages, as well as to control the movements of the drone in a simulated domestic environment. 
The conducted experiments involve participants giving voice commands to the drone in both languages in order to compare the effectiveness of each of them, considering the mother tongue of the participants in the experiment. 
Additionally, different levels of distortion have been applied to the voice commands in order to test the proposed approach when facing noisy input signals. 
The obtained results show that the unmanned aerial vehicle is capable of interpreting user voice instructions achieving an improvement in speech-to-action recognition for both languages when using phoneme matching in comparison to only using the cloud-based algorithm without domain-based instructions.
Using raw audio inputs, the cloud-based approach achieves 74.81\% and 97.04\% accuracy for English and Spanish instructions respectively, whereas using our phoneme matching approach the results are improved achieving 93.33\% and 100.00\% accuracy for English and Spanish languages.
\end{abstract}

\thispagestyle{fancy}


\section{Introduction} 
\begin{figure}[t]
    \centering
    \subfigure[DJI Phantom 4 radio controlled quadrotor operating in a real environment. 
    ]{\includegraphics[width=0.4\columnwidth]{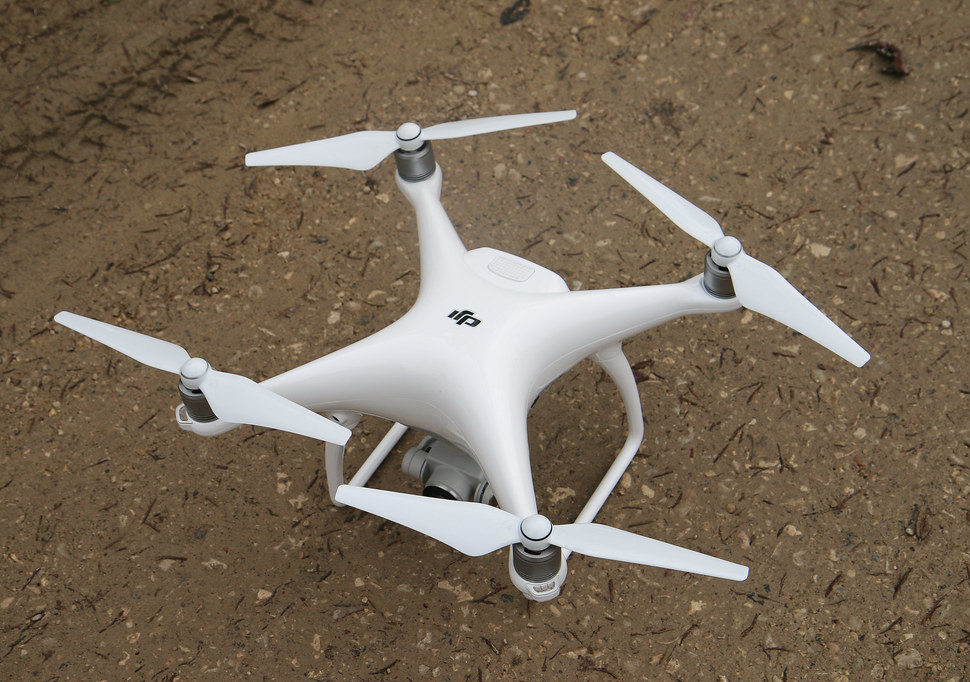}
        \label{fig:drone}
    }\hspace{1.5cm}
    \subfigure[Simulated quadrotor drone in V-REP used in the proposed domestic scenario.
    ]{\includegraphics[width=0.4\columnwidth]{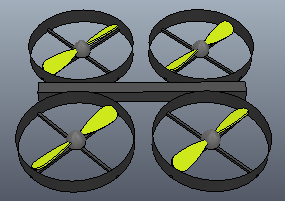}
        \label{fig:vrep}
    }
    \caption{Example of quadrotor drones. The unmanned aerial vehicles are shown in both real-world and simulated environments.}
    \label{fig:DroneComparison}
\end{figure}

Presently, unmanned aerial vehicles (UAVs) are more frequently used with a wide variety of applications in areas such as security, industry, food, and transport, among others~\cite{c1}.
In this regard, it is essential to incorporate solutions that provide UAVs with the ability to be controlled remotely, making feasible the communication of the orders that must be executed.
A very popular kind of UAV is a drone, which is a mobile robotic structure capable of flying that may be operated remotely.
There are several types of UAVs according to their properties of frame, propellers, engine, system of power, electronic control, and communication system~\cite{c1}.
Commonly, UAVs have a built-in camera to capture video from its flight; others additionally include a thermal infrared camera~\cite{seymour2017automated} to capture the wildlife without disturbing its environment, and other UAVs include a radio frequency sensor to detect hazardous waste in railway accidents~\cite{imp2}.
Moreover, UAVs have become helpful vehicles for the acquisition of data as well as for the transport of elements with no human presence.
An example of a quadrotor drone that can be operated in real-world scenarios by radio control is shown in Figure~\ref{fig:drone}.

A significant part of the UAV functionalities and advantages lies in the sensors~\cite{c1} that provide an extension to its capacities in order to obtain information about the environment in where it is deployed.
Nevertheless, there are just a few add-ons focused on extending its remote control capability.
For instance, this problem has been addressed by Fernandez et al.~\cite{fernandez2016natural} where voice control is proposed using dictionaries.
The proposed technique comprises 15 commands for UAV control in a given language.

In this paper, we present an experimental approach for drone control through a cloud-based speech recognition system, improved by a domain-based language.
The cloud-based automatic speech recognition is carried out using the \textit{Google Cloud Speech}\footnote{https://cloud.google.com/speech-to-text/} (GCS) service~\cite{schalkwyk2010your} through the Web Speech API\footnote{https://wicg.github.io/speech-api/}~\cite{adorf2013web}, which in this context has been customized with a domain-based dictionary for the proposed scenario as proposed in~\cite{Twiefel14}.
Therefore, we combine GCS and a predefined language based on the domain of the problem to convert the voice into text, which is then interpreted as an instruction for the drone.
Our domain-based language is a dictionary comprising 48 instructions, some of which are in Spanish and others in English.
These commands are interpreted and mapped into one of the nine available actions that the drone can execute. 
Experiments are performed in a simulated domestic environment that includes chairs, tables, and shelves, among others.
This novel approach contributes to the state of the art by improving the automatic speech recognition, in terms of action classification for drone control, by scaffolding the raw voice input signal through domain-based phoneme matching.
Additionally, the proposed approach contributes by allowing an enhanced drone control independently of the user's native language.
In this regard, even in the presence of a noisy signal, defective English instruction utterances do not significantly affect the speech recognition system after phoneme matching, achieving in all tested cases high rates of accuracy.

During the experimental setup, participants gave voice commands to the drone in a simulated environment. 
In general, Spanish language instructions were better understood than English language instructions, mainly due to the native language spoken by the participants, however, in both cases, the success rate for recognition of the instructions was improved by using domain-based instructions.
The implementation of the simulated scenario, including the drone and the speech recognition features mentioned above, is carried out using the V-REP\footnote{\label{note:vrep}From November 2019, V-REP simulator has been replaced by CoppeliaSim version 4.0. See https://www.coppeliarobotics.com/} robot simulator, which has been previously used in domestic robotic scenarios~\cite{cruz2016learning, cruz2017agent, cruz2018action, moreira2020deep, cruz2020explainable}. 
V-REP~\cite{rohmer13vrep} is a simulation tool that allows experimentation with virtual robots, provides a graphical interface for creating and editing different simulated models, as well as designing the environment with the necessary elements.
Figure~\ref{fig:vrep} shows the simulated drone that is used within the home environment implemented in this work.

\section{Related Works}

\subsection{Unmanned Aerial Vehicle Control}
For some years now, unmanned aerial vehicles (UAVs), such as drones, have been more in demand on the market~\cite{ventadrones}. 
Nevertheless, UAVs were created many decades ago, when Archibald Low proposed the first drone in 1916 on a project for the British Air Ministry to develop an unmanned defense aircraft to be used against German airships~\cite{marshall2016UAVintroduction}. 
From Low's development of the first radio control UAV to the current drone industry, a plethora of changes have occurred.
Nowadays, drones play an active role in many areas, such as military, agriculture, recreation, and search and rescue~\cite{c1}.
The massive emergence of UAVs and their extensive applications in different fields have led to the development of simpler control forms for non-expert end-users.
Implementation of UAV systems has been proven to be cost-effective in covering vast area extensions, e.g., for data acquisition tasks~\cite{muchiri2016review}.
Moreover, drones present higher maneuverability in areas where other traditional unmanned air vehicles have shown inefficacy~\cite{amin2016review}.

Since UAVs are designed to fly with no onboard pilot, a self-driving UAV is indeed a desirable characteristic to be present in these vehicles. 
Self-controlled drones have achieved a high autonomous flight level, as specified in \cite{clough2002metrics}, addressed by novelty machine learning techniques \cite{Peng2012OnlineControl, al2018design, ivanovas2018block, pham2018autonomous, Shiri2020uav, kusyk2020artificial}.
Many of these works have achieved successful performances in online path planning, however, the results still rely on previously presented paths, or in a route tracking pattern.
In this regard, when we refer to an autonomous system, it does not mean it is necessarily an intelligent system.
The main challenge to enable intelligent UAVs is the automated decision-making process as pointed out in \cite{Chen2009Survey}.
Other authors have addressed UAV online control as a semi-autonomous system \cite{Quigley2004SemiAuto} capable of being controlled externally through a hardware interface.
The semi-autonomous UAV control technique allows a human operator to intervene in the drone's actions when required to make paths that are more precise during flights \cite{Wopereis2015SemiAuto}.
Moreover, related task-specific works have been proposed aiming to use the UAV as an aiding tool for the users' tasks \cite{PerezGrau2017SemiAuto, Imdoukh2017SemiAuto}.
The use of semi-autonomous remotely operated drones for different tasks lets the users make corrections in the flight path during the mission, allowing them to obtain overall improved results for many goal-specific purpose systems.

A simple UAV controller aims to get familiar with the use of the drone to all possible users of many different fields.
For instance, in Fernandez et al. \cite{fernandez2016natural}, the authors experimented with several natural user interfaces for human-drone interaction, among them the gesture and speech control.
For gestures, experiments included body and hand interaction.
Each interface was tested with different users in public areas, achieving overall positive comments.
A more recent gesture-based control of semi-autonomous vehicles \cite{sanders2020design}, tested in virtual reality and real-world environments, has shown that users in some situations still prefer to use a joystick device to control the vehicle, mainly because they are more habituated to such technology.
However, the authors also conclude that hand gesture-based control is more intuitive and easy for users to learn how to drive the vehicle quickly. 
Although the actual technology can capture a wide range of hand motions with high precision, humans may not be so precise in all circumstances, therefore, a control system may benefit from supplementary methods to preprocess the inputs and smooth the movements.
In terms of speech for drone's control, Wuth et al. \cite{wuth2020role} demonstrated that users find speech-based communication to be transparent and efficient.
This efficient communication relies on the knowledge of the context for the specific task addressed by the drone.
Overall, the context plays an important role in human-drone environments to achieve effective control.

\subsection{Speech Control}

Currently, some works have addressed UAV control using more natural interfaces for people, e.g., through automatic speech recognition.
For instance, Lavrynenko et al.~\cite{lavrynenko2016voice} propose a radio-based remote control system, in which a semantic identification method based on mel-frequency cepstral (MFC) coefficients is used.
The captured audio is translated to an action that, in turn, is transferred to the UAV for its execution.
Each time a microphone captures a new voice command, the system computes the cepstral coefficient.
The coefficient is compared against a database of cepstral coefficients using the minimum distance criterion to match the desired command.
Nevertheless, the database of cepstral coefficients only comprises four voice commands, corresponding to each direction of the UAV. 

A more accurate speech recognition method was presented by Fayjie et al.~\cite{fayjie2017drone}.
In this work, the authors utilized a hidden Markov model for speech recognition with voice adaptation in order to control the UAV.
Their proposal was based on a speech-decoding engine called Pocketsphinx, used with ROS in the Gazebo simulator.
The speech decoding worked with the CMU Sphinx Knowledge Base Tool, settled with seven actions to control altitude, direction, yaw, and landing.
Unfortunately, the CMU Sphinx Knowledge Base Tool is not being actively developed and is considered deprecated against modern neural network-based approaches.
Another similar approach was addressed by Landau et al.~\cite{landau2017drone}, where the authors used the Nuance speech recognition service.
They proposed a hands-free UAV control with voice commands, to actuate over a DJI Phantom 4 drone, developed with the DJI Mobile SDK for iOS.
The proposed architecture was composed of a Bluetooth hands-free for voice capture and the speech commands were translated and evaluated using regular expressions.
The regular expressions were divided into three groups, the first group contained possible words to move the drone in any direction, the second group contained possible words to also move the drone in any direction but with an established amount of distance, and the third group contained words to take-off and land the drone.
The implementation of this work is limited to be used through a Bluetooth hands-free device connected to an Android smartphone in order to control only DJI manufactured drones.

Chandarana et al.~\cite{chandarana2017uav} presented a custom-developed software using speech and gesture recognition for UAV path planning.
In their work, the authors performed a comparison of natural language interfaces using a mouse-based interface as a baseline and evaluating individual users who have to complete a flight path composed of flight trajectories or actions.
The authors proposed a software where the users interact using either a mouse, gesture or speech, in order to build three specific flight paths.
The speech recognition phase was handled using the CMU Sphinx 4-5prealpha speech-to-text software, used with rule-based grammar, allowing the system to hear compound names formation, e.g., forward-left and backward-right trajectories, among others.
Their work also presented an evaluation of the user's response to natural language interfaces.
Although the best performance was achieved using the mouse-based interface, users reported preference in using speech for mission planning.

Additionally, a multi-modal approach considering voice interaction with drones was developed in~\cite{fernandez2016natural}. 
In this work, a word dictionary for speech recognition was utilized, however, only 15 different commands in one language were allowed to control the UAV.
In~\cite{Quigley2004SemiAuto}, a speech controller was developed to recognize commands sent to a semi-autonomous UAV. 
In the experiments, flight tests were conducted revealing that ambient noise and conversation could considerably affect the reliability of the speech recognition system. 
In~\cite{jones2010towards}, a study of gesture and speech interfaces for interaction with drones in simulated environments was conducted.
It was shown that participant subjects generally preferred to use lower-level commands, such as left or right to control the drone. 

One of the most recent works is an extension of~\cite{lavrynenko2016voice} developed by Lavrynenko et al.~\cite{lavrynenko2019voice}.
In this extension, the authors present a similar radio-based control system with cepstral analysis, however, in this work they also add encrypted communication.
The proposed architecture uses a voice-control panel that handles the encryption, including the cepstral and wavelet coefficients.
Moreover, the inverse process of coefficient quantization is performed, comparing it to the cepstral database using the minimum distance criterion.
Both parts, the encryption and decryption, present an encryption key, which works with signal filters acquiring the features of the speech.

\section{Proposed Architecture}
In the last decade, a technology actively developed has been the Natural User Interface (NUI)~\cite{lopez2017alexa}, which focuses on communication in a natural way between digital devices and end-users.
In these approaches, it is fostered that humans can interact with a machine in such a way that there is no control device in order to generate input information to the machine, such as keyboards or touch screens, or any other device that people must have physical contact ~\cite{glonek2012natural}.
There are different types of NUIs, and some of them include:

\begin{itemize}
    \item \textbf{Voice recognition}: The interface must be able to recognize instructions through the user's voice. 
    \item \textbf{Gesture recognition}: The interface can capture gestures from the human body and interpret them.
    \item \textbf{Visual marker interaction}: Visual markers are added, which are captured by a camera and recognized by the machine.
\end{itemize}


In the proposed architecture, the drone interaction is not carried out through remote control, but rather it makes use of a Natural-language User Interface (NLUI), interpreting instructions from humans through automatic speech recognition. 
For instruction interpretation, the person's voice is captured by a microphone connected to a computer that executes the algorithms to process the received audio signal. 
The microphone can be either the built-in one from a laptop computer or any other external device, however, the quality of the signal captured may considerably vary and, in turn, impact the accuracy of the interpretation~\cite{cruz2015interactive}. 
In order to transform the audio signal into text, Google Cloud Speech (GCS) is used in combination with a domain-based language.
Audio streams are received from the microphone and sent to the cloud-based GCS service through the Web Speech API, from where we obtain a recognized sentence as a hypothesis.
Following, the hypothesis is compared to our domain-based dictionary performing a phoneme matching using the Levenshtein distance~\cite{levenshtein1966binary}.

\begin{figure}[t]
    \centering
    \includegraphics[width=0.7\columnwidth]{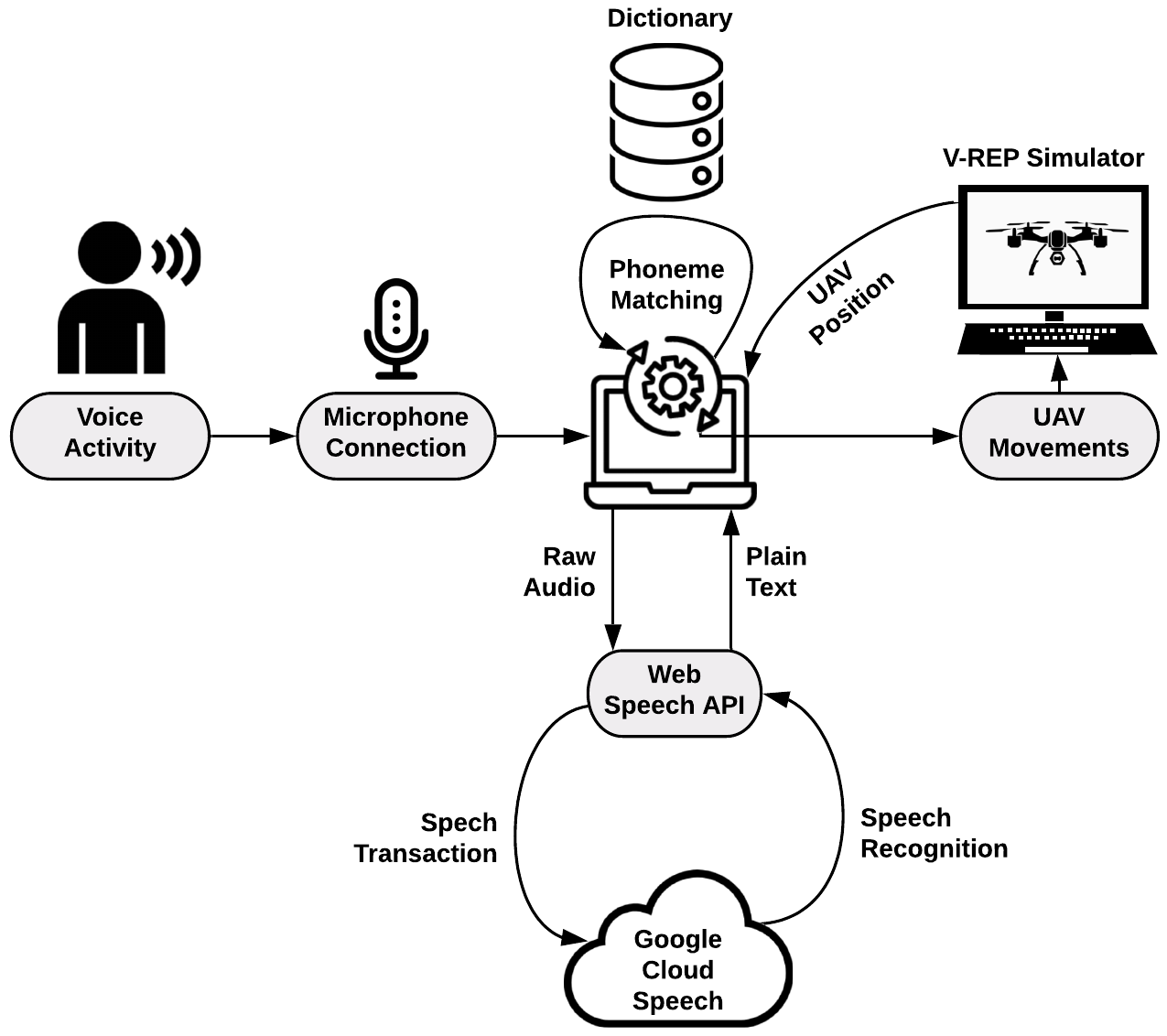}
    \caption{The proposed architecture for UAV control through speech.
    In this representation, a person speaks the instructions in a microphone and these are processed by our algorithm.
    The instruction is then classified using the domain-based dictionary and executes for the UAV.}
    \label{fig:ordenes}
\end{figure}

The Levenshtein distance $\mathcal{L}$, also known as the edition distance, is the minimal amount of operations needed to transform a sentence $s_x$ into another sentence $s_y$. 
The comparison is performed by comparing the characters inside $s_x$ to the ones inside $s_y$.
The operations considered to transform the sentence comprise substitutions, insertions, and deletions.
The cost of each edition operation is equal to 1. 
The distance is computed recursively as $\mathcal{L}_{s_x, s_y}(|s_x|, |s_y|)$ with $|s_x|$ and $|s_y|$ being the length of the sentences $s_x$ and $s_y$ respectively, and where the $i$-th segment of the sentence is computed as shown in Equation \eqref{Eq:Levenshtein}. 
In the equation, $c_{s_{x_i},s_{y_j}}$ is 0 if $s_{x_i}=s_{y_j}$ and 1 otherwise.
Thus, the cost of transforming the sentence $s_1=$ "to the left" to $s_2=$ "go to the left" is equal to $\mathcal{L}_{s_1, s_2}=3$ since involves the insertion of 3 new characters. 
Likewise, the cost of transforming the sentence $s_3=$ "go right" to $s_4=$ "go left" is equal to $\mathcal{L}_{s_3, s_4}=4$ since the number of operations needed is 3 substitutions and 1 deletion.

\begin{equation}
    \mathcal{L}_{s_x, s_y}(i, j) = \left\{
    \begin{array}{l l}
         \textrm{max}(i, j) & \textrm{if min} (i, j) = 0 \\
    \textrm{min} \left\{
    \begin{array}{l}
         \mathcal{L}_{s_x, s_y}(i - 1, j) + 1  \\
         \mathcal{L}_{s_x, s_y}(i, j - 1) + 1  \\
         \mathcal{L}_{s_x, s_y}(i - 1, j - 1) + c_{s_{x_i},s_{y_j}} \\ 
    \end{array}
    \right. & \textrm{if min} (i, j) \neq 0\\
    \end{array}
    \right.
\label{Eq:Levenshtein}    
\end{equation}

To perform the phoneme matching, the Levenshtein distance is computed between the recognized hypothesis and the domain-based dictionary. 
Afterward, the instruction showing the minimum distance is selected. 
Once the voice command is converted into text, the signal can be processed and classified as an instruction for the UAV, which in our scenario is a drone within the V-REP robot simulator. 
Figure~\ref{fig:ordenes} shows the proposed architecture.
Moreover, Algorithm~\ref{alg:algoritmo} shows the operations to carry out the control of the drone through voice commands considering both with and without phoneme matching.

\begin{algorithm}[t]
	\caption{Algorithm implemented for the interpretation of an audio signal into an instruction for the drone. The algorithm comprises two sections for speech recognition with and without phoneme matching.}
	\label{alg:algoritmo}
	
	\begin{algorithmic}[1]
	    \State \textbf{Initialize} dictionary $D$ with instructions $i$ and classes $C$.
    	\Repeat
        	\State \textbf{Wait} for microphone \textit{audio} signal.
        	\State \textbf{Send} \textit{audio} signal to Google Cloud Speech.
            \State \textbf{Receive} \textit{hypothesis} $h$.
            \If{phoneme matching is activated}
                \For{each instruction $i \in D$}
                    \State \textbf{Compute} $\mathcal{L}_{h, i}$.
                \EndFor
                \State \textbf{Choose} instruction as $\min \mathcal{L}_{h, D_i}$. 
                \State \textbf{Match} chosen instruction to action class $a \in C$.
                \State \textbf{Execute} action class $a \in C$ in the scenario.
            \Else
                \For{each instruction $i \in D$}
                    \State \textbf{Compare} $h$ to $D_i$.
                    \If{$h \in D$}
                        \State \textbf{Match} instruction $i \in D$ to action class $a \in C$.
                        \State \textbf{Execute} action class $a \in C$ in the scenario.
                        \State \textbf{Exit} loop.
                    \EndIf
                \EndFor
            \EndIf
        \Until{an exit instruction is given}
	\end{algorithmic}
\end{algorithm}


\section{Experimental Setup}

\begin{figure}[t]
    \centering
    \includegraphics[width=0.65\columnwidth]{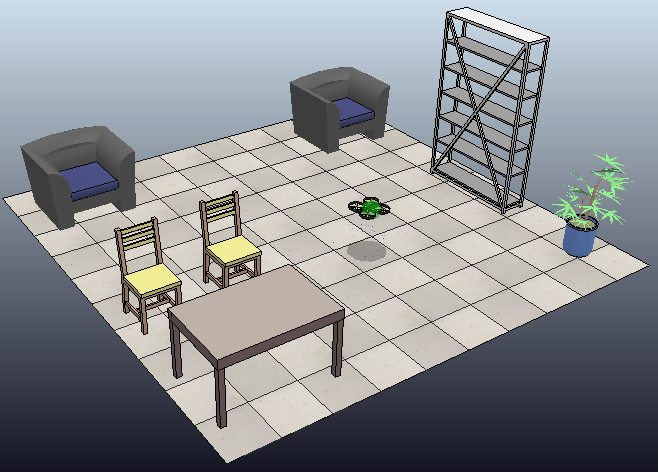}
    \caption{The simulated domestic environment in V-REP with a quadrotor and daily-used furniture such as sofas, chairs, a table, a shelve, and a plant.}
    \label{fig:simulacion}
\end{figure}

Different tools have been used for the development of this project. 
One of them is V-REP~\cite{rohmer13vrep}, a closed-source simulation software freely available with an educational license for several operating systems, such as Linux, Windows, and iOS, for simulating different types of robots in realistic environments. 
Additionally, it has a wide range of API libraries to communicate the simulator with different programming languages~\cite{ayala2020comparison}. 
For this project, a simulated scenario has been built comprising a series of daily-used furniture in a domestic environment as well as the simulated UAV. 
We make use of the flight stabilization controller provided by the simulator in order to keep the focus of the work on the execution of commands through voice directions. 
The experimental scenario can be seen in Figure~\ref{fig:simulacion}.

In our scenario, once an instruction is given to the drone, it is executed continuously until another action is instructed.
Therefore, to stop the vehicle, it is necessary to instruct explicitly the action "stop".
The only exception to the previous rule is the execution of the action "down", which may be automatically stopped in case the drone reaches 0.5m distance from the ground, in such a case, the movement is ceased to avoid the collision.
All possible actions are shown in Table~\ref{tab:instrucciones}, where it is possible to see the nine action classes defined in the simulated scenario.
The instructions can be given using both languages Spanish and English.

\begin{table}[t]
    \caption{Description of allowed commands to produce an action to control the UAV.}
    \centering
    \begin{tabular}{cll}
        \hline
        \textbf{Nr.} & \textbf{Action classes} & \textbf{Description}\\
        \hline
        1  & Up           & Increase the UAV's altitude  \\
        2  & Down         & Decrease the UAV's altitude \\
        3  & Go right     & Move the UAV to the right\\
        4  & Go left      & Move the UAV to the left\\
        5  & Go forward   & Move the UAV forward\\
        6  & Go back      & Move the UAV backward\\
        7  & Turn right   & Turn the UAV 90$^{\circ}$ clockwise\\
        8  & Turn left    & Turn the UAV 90$^{\circ}$ counterclockwise\\
        9  & Stop         & Stop the UAV\\ 
        \hline
     \end{tabular}
     \label{tab:instrucciones}
\end{table}

For the architecture implementation, the programming language Python is used and connected to the simulator through the V-REP API, in order to pass the instructions between the automatic speech recognition algorithm and the simulator.
As mentioned, words and phrases may be uttered by the users in two languages, Spanish and English.
The selection and benefits of using these languages are twofold.
On the one hand, the mother tongue of participants in the experimental stage is Spanish, and, on the other hand, the global use of English language, therefore, a comparison of accomplished accuracies is carried out using both languages.

Each action class has more than one way to be described to execute a movement, e.g., the action "down" can be instructed by saying the word "baja" or the sentence "disminuir altura" in Spanish, or also in the form "go down" or simply "down" in English. 
It is important to note that not all phrases are necessarily correct in grammar terms either in Spanish or in English.
The reason is that we are not assuming here that an end-user gives all the time an instruction using fully correct sentences.
It is known that spoken language on many occasions is less structured and, therefore, lacks formality not following grammar rules. 
In this regard, we define a domain-based dictionary comprising 48 sentences belonging to the nine action classes. 
It is important to note that the classes “go” and “turn” differentiate since the former moves the drone to the left or right in x,y coordinates keeping the drone's orientation, and the latter changes the drone's yaw angle by 90° clockwise or counterclockwise.

The experiments were run in a computer with the following characteristics: Intel Core i7-8750H processor, 8GB DDR4 2666MHz RAM, NVIDIA GeForce GTX 1050Ti with 4GB of GDDR5, and Windows 10 Home. 
The Internet connection used was an optical fiber with a 300/100 Mbps download/upload speed.

\section{Results}


In this section, we show the main obtained results by testing the proposed algorithm. 
In our experiments, apart from testing with online instructions uttered by different people, recordings from different locations are also used, such as open spaces, offices, and classrooms. 
Recordings present an averaged signal to noise ratio (SNR) of \num{-3,09E-04} dB, showing a slightly better ratio for sentences in Spanish, which may also be attributed to the native language of the participants. 
The SNR values are shown in Table \ref{tab:srn} for each action class in both languages. 

\begin{table}[H]
    \caption{Raw input SNRs (dB) for each action class in both Spanish and English language.}
    \centering
    \begin{tabular}{lccc}
        \hline
        \textbf{Class} & \textbf{English} & \textbf{Spanish} & \textbf{Average}\\
        \hline

        Up         & \num{-4,21E-04} & \num{5,33E-04}  & \num{5,57E-05}  \\
        Down       & \num{-9,06E-04} & \num{-2,37E-05} & \num{-4,65E-04} \\
        Go Right   & \num{-6,93E-04} & \num{-2,09E-04} & \num{-4,51E-04} \\
        Go Left    & \num{-8,03E-04} & \num{-3,16E-05} & \num{-4,18E-04} \\
        Go Forward & \num{-3,85E-04} & \num{-1,36E-04} & \num{-2,60E-04} \\
        Go Back     & \num{-6,86E-04} & \num{-9,70E-06} & \num{-3,48E-04} \\
        Turn Left  & \num{-9,54E-04} & \num{-5,55E-05} & \num{-5,05E-04} \\
        Turn Right & \num{-7,79E-04} & \num{2,68E-04}  & \num{-2,55E-04} \\
        Stop       & \num{-2,94E-04} & \num{3,02E-05}  & \num{-1,32E-04} \\ 
        \hline
        \textbf{Average}   & \num{-6,58E-04}  &  \num{4,06E-05}  &\num{-3,09E-04} \\
        \hline
     \end{tabular}
     \label{tab:srn}
\end{table}

In order to determine the accuracy of the proposed algorithm, tests in two languages with and without phoneme matching have been executed using three different setups, i.e., raw input, 5\% noisy input, and 15\% noisy input. 
The two noisy setups are added to test the robustness of the algorithm in presence of noise and include uniform noise $n_1 = 0.05$ (uniformly distributed $U(-n_1,n_1)$) and $n_2 = 0.15$ (uniformly distributed $U(-n_2,n_2)$) equivalent to 5\% and 15\% with respect to the original raw input. 
For each setup, each action class is performed 15 times for each language, therefore, each class is called a total of 30 times, 15 for English and 15 for Spanish.
Overall, 270 instructions are tested for each setup, 135 for each language. 
A total of 5 people participated in this experimental test. 
Although we are aware that the number of participants is rather small, we are still able to draw significant conclusions to outline future experiments.
Additionally, this work includes people from different age segments, going from 19 years old to 56 years old (mean $M=35.4$, standard deviation $SD=18.45$, 3 women, 2 men). 

\begin{figure}[t]
  \centering
  \subfigure[Raw input, no phoneme matching.] {\includegraphics[width=0.3\textwidth]{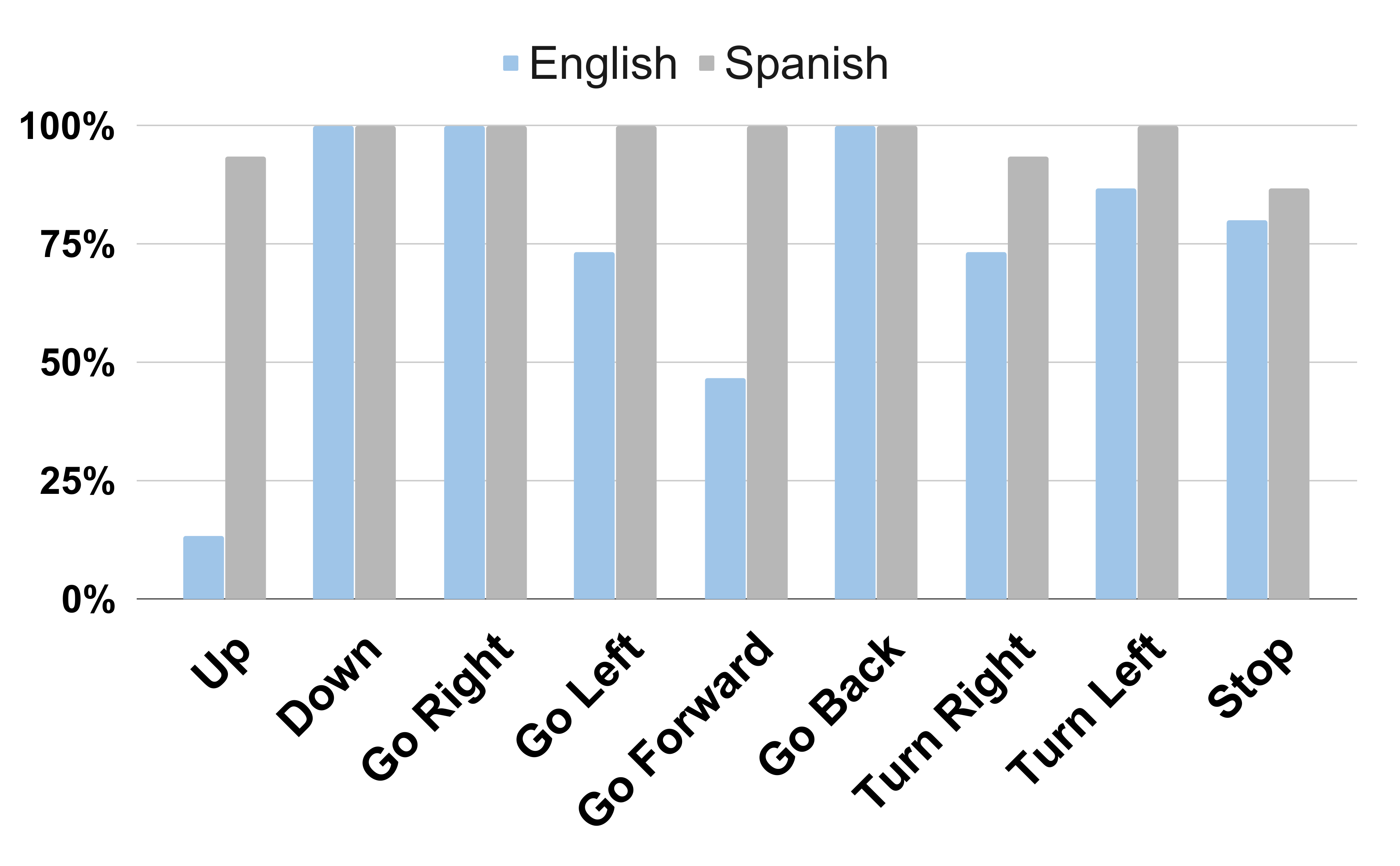} \label{fig:resultadosNPM}
  }\hfill
  \subfigure[Noise 5\%, no phoneme matching.] {\includegraphics[width=0.3\textwidth]{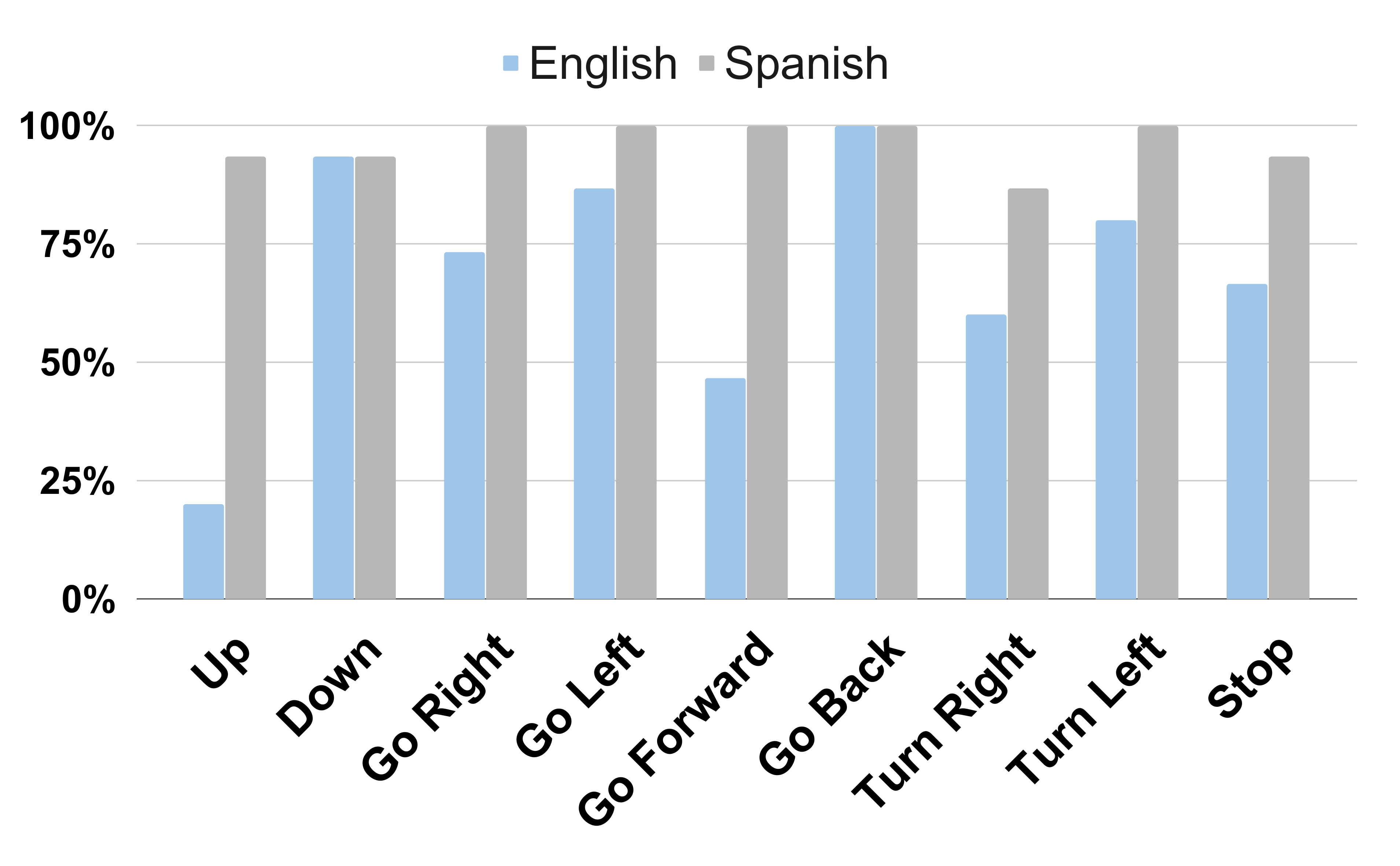} \label{fig:resultadosNPM5}
  }\hfill
  \subfigure[Noise 15\%, no phoneme matching.] {\includegraphics[width=0.3\textwidth]{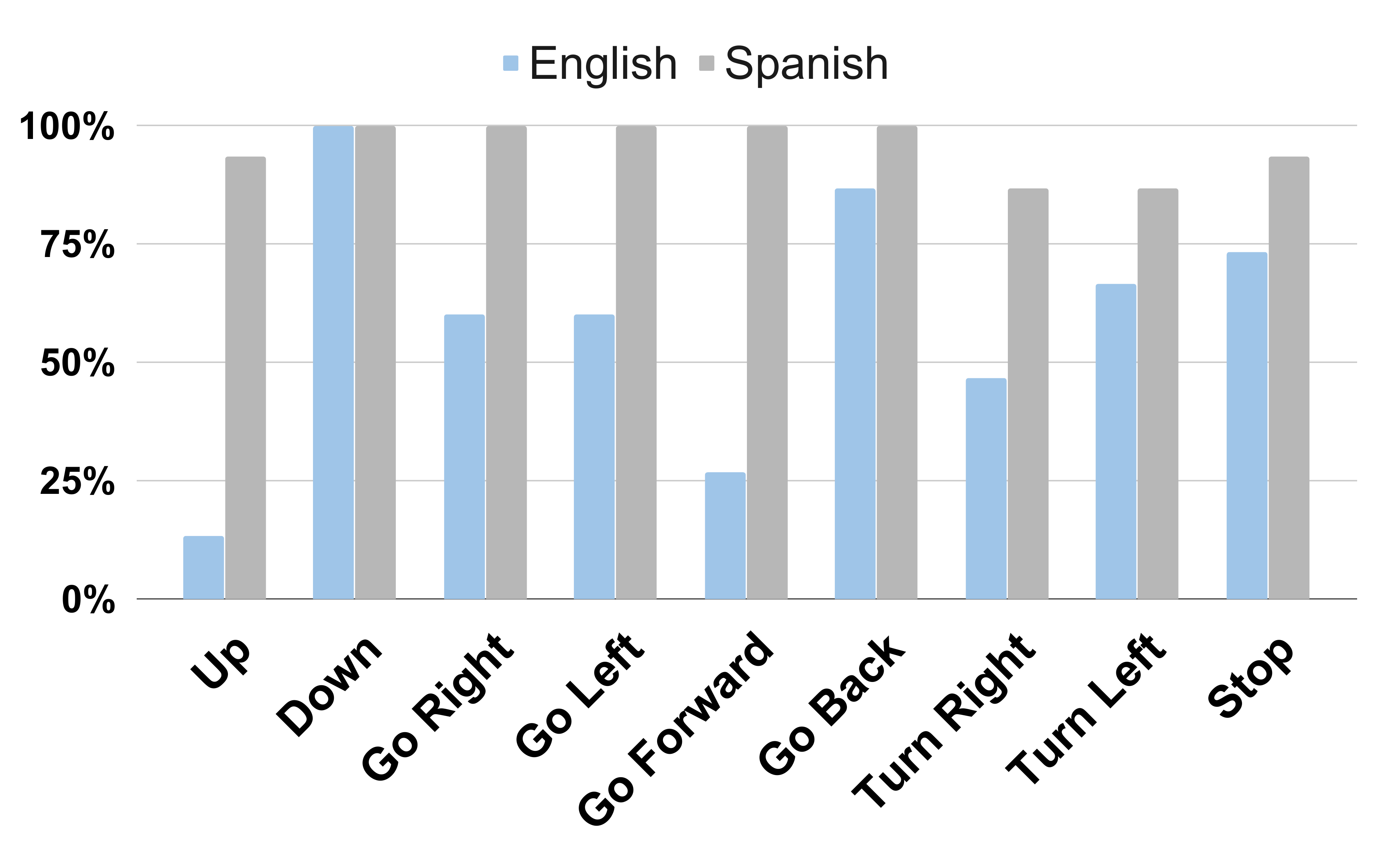} \label{fig:resultadosNPM15}
  }
  
  \subfigure[Raw input, with phoneme matching.] {\includegraphics[width=0.3\textwidth]{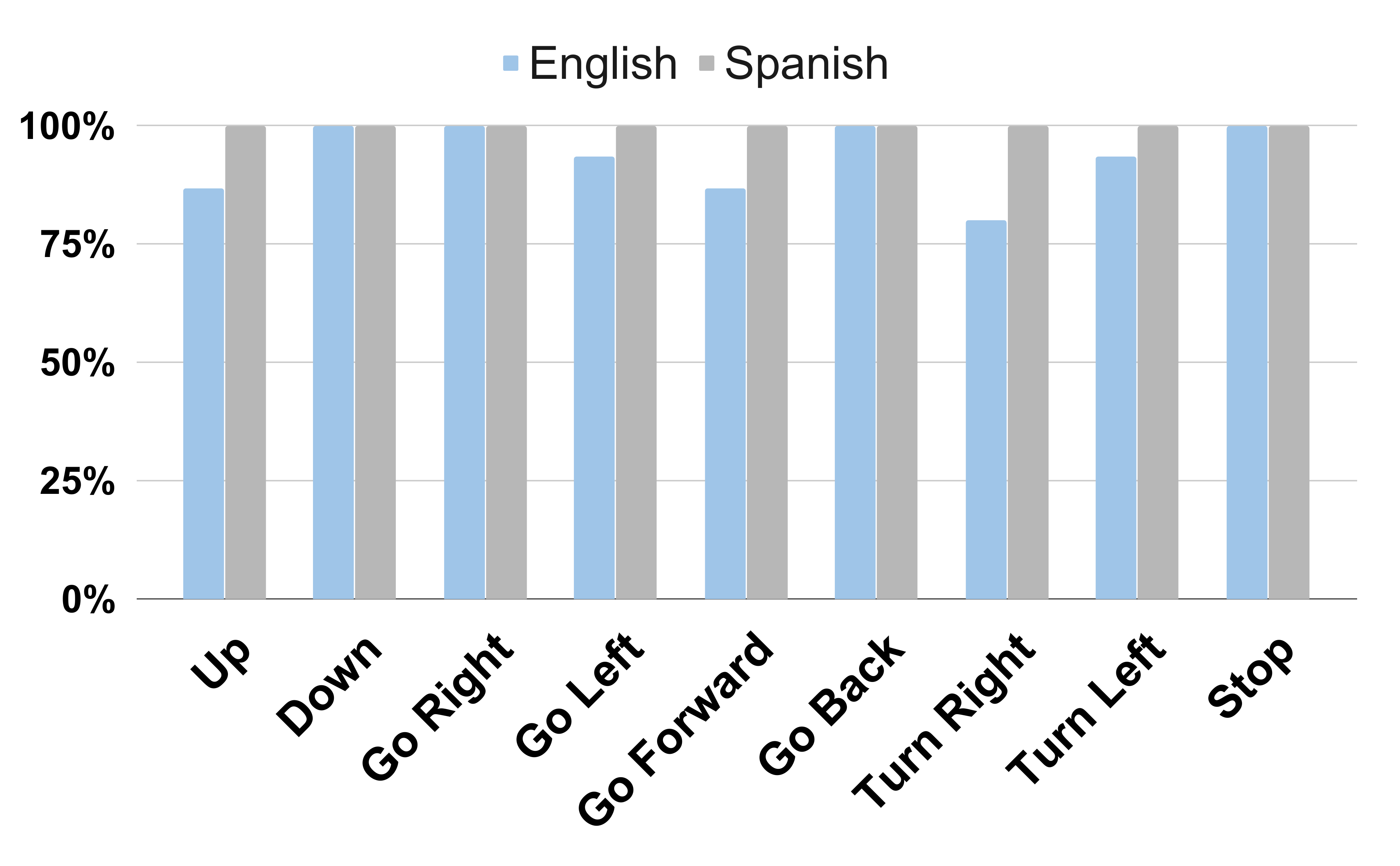} \label{fig:resultadosWPM}
  }\hfill
  \subfigure[Noise 5\%, with phoneme matching.] {\includegraphics[width=0.3\textwidth]{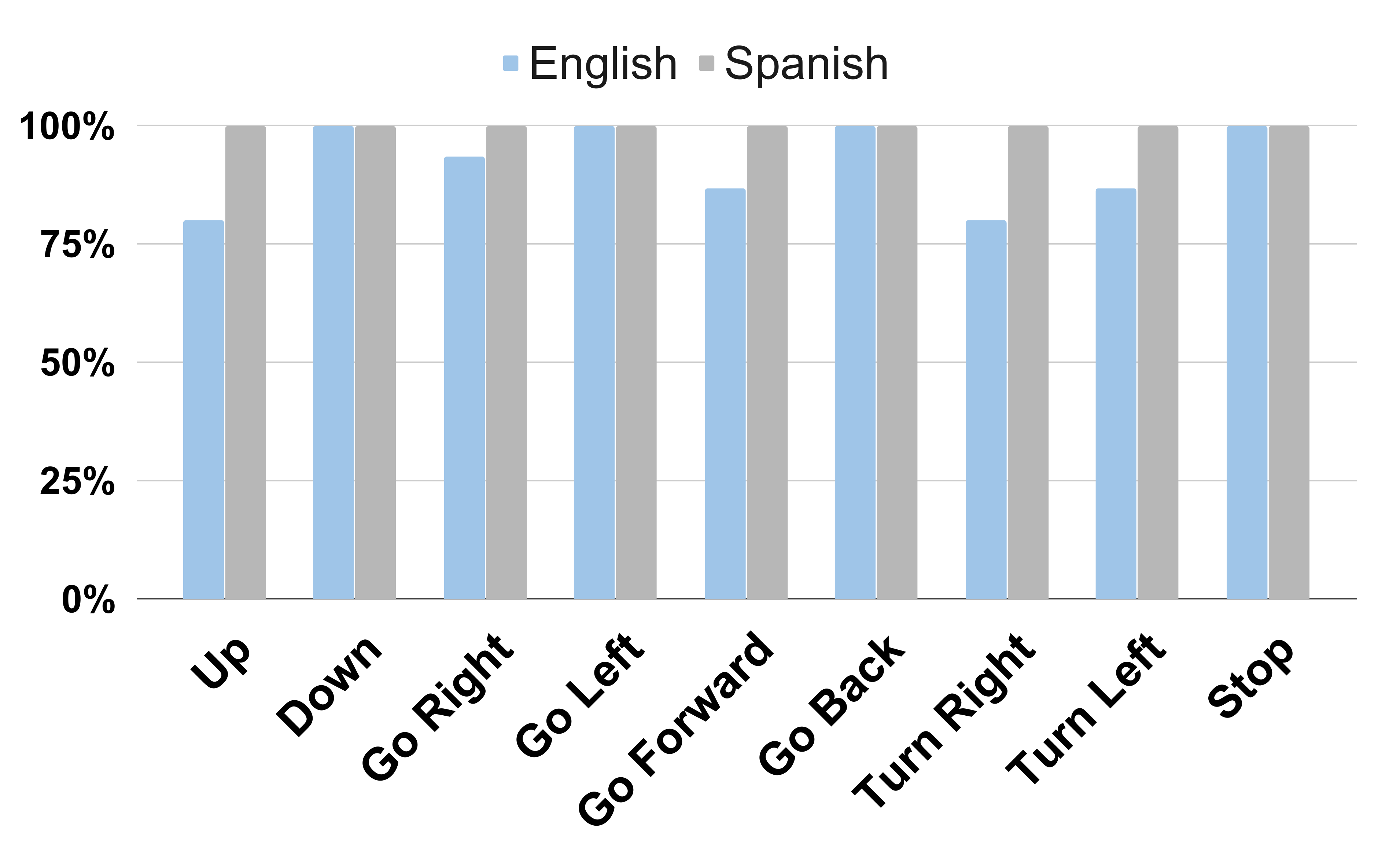} \label{fig:resultadosWPM5}
  }\hfill
  \subfigure[Noise 15\%, with phoneme matching.] {\includegraphics[width=0.3\textwidth]{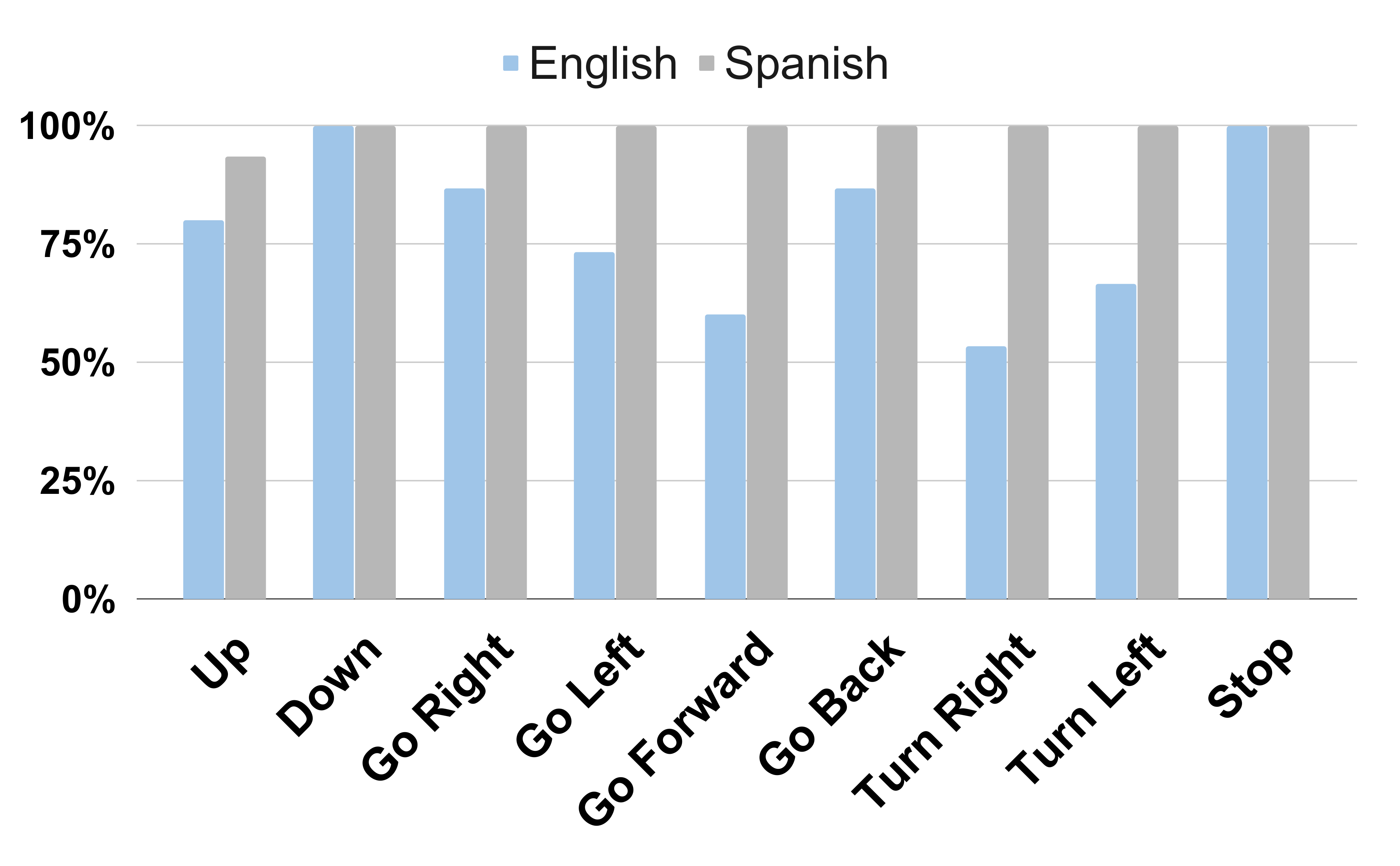}
  \label{fig:resultadosWPM15}
  }
  \caption{Average recognition accuracy for each action class in Spanish and English languages with different levels of noise in the input signal. 
  Without using phoneme matching, the text received from the cloud-based service is directly transferred to the scenario.
  This implementation shows a considerable difference between languages due to the user's native language.
  Using phoneme matching, the text received from the cloud-based service is compared to the instructions within the domain-based dictionary.
  The use of phoneme matching shows an improvement in speech-to-action recognition for both languages, decreasing the difference of accuracy between them. 
  }
  \label{fig:OverallResultados}
\end{figure}

Figure \ref{fig:OverallResultados} shows the obtained accuracy using English and Spanish instructions for all the levels of noise.
Figures \ref{fig:resultadosNPM}, \ref{fig:resultadosNPM5}, and \ref{fig:resultadosNPM15} show the accuracy without using phoneme matching, i.e., the algorithm compares the text received from GCS directly to our domain-based dictionary trying to find an exact coincidence; otherwise, it is not recognized or labeled as “no class”.
When phoneme matching is not used, a considerable accuracy difference can be noticed between Spanish and English commands, with the former presenting the highest recognition values. 
In this regard, the users instructing in Spanish, i.e., their native language, achieve better action recognition in comparison to English commands, likely related to the poor utterance of the words in a foreign language.
Figures \ref{fig:resultadosWPM}, \ref{fig:resultadosWPM5}, and \ref{fig:resultadosWPM15} show the obtained recognition accuracy using the domain-based language for phoneme matching.
When using phoneme matching, the difference in the achieved recognition between both languages is attenuated by our algorithm which looks for the most similar instruction to classify the audio input. 

In terms of noisy inputs, as mentioned, we have performed experiments using the raw input, 5\% noisy input, and 15\% noisy input.
Figures \ref{fig:resultadosNPM} and \ref{fig:resultadosWPM} show the obtained results without and with phoneme matching technique when using the raw audio input. 
When no phoneme matching is applied, the algorithm recognized 232 out of 270 instructions considering both languages, achieving 85.93\% accuracy in voice-to-action recognition. 
Particularly, the use of Spanish language achieves 97.04\% accuracy, while the use of English reaches 74.81\% accuracy.
However, when phoneme matching is used the algorithm considerably improves the recognition accuracy for both languages achieving 96.67\% accuracy. 
While the use of Spanish language achieves 100.00\% accuracy, the recognition of English commands significantly improves in comparison to the non-phoneme-matching approach, reaching 93.33\% accuracy.  

Following, in order to test the robustness of the proposed method, we applied a 5\% of noise to the audio input.
The obtained results without and with phoneme matching can be seen in Figures \ref{fig:resultadosNPM5} and \ref{fig:resultadosWPM5} respectively.
On average without applying phoneme matching, the algorithm recognized 214 out of 270 instructions considering both languages, achieving an accuracy of 82.96\% in voice-to-action recognition. 
In particular, Spanish instructions achieve 96.30\% accuracy, while English instructions 69.63\% accuracy.
Using phoneme matching, the algorithm accomplished 95.93\% accuracy, i.e., 100.00\% accuracy for Spanish commands and 91.85\% accuracy for English commands.
When comparing the recognition accuracy using a 5\% noisy input to the use of the raw input, the obtained results just slightly worsen, especially when phoneme matching is used, showing the robustness of the proposed approach in presence of noisy audio inputs.  

Finally, we used an audio input signal with 15\% of noise.
The obtained results are shown in Figures \ref{fig:resultadosNPM15} and \ref{fig:resultadosWPM15} without and with phoneme matching respectively.
Without applying phoneme matching, the algorithm recognized 198 out of 270 instructions, achieving 77.41\% accuracy in speech-to-action recognition on average for both languages. 
The use of Spanish instructions accomplished 95.56\% accuracy, while English instruction 59.26\% accuracy. 
Introducing phoneme matching in this setup, the algorithm accomplished 88.89\% accuracy, i.e., 99.26\% accuracy for Spanish instructions and 78.52\% accuracy for English instructions.
Although like in the previous case, the introduction of noise did affect the obtained recognition accuracy, which was expected due to the input signal distortion, the use of phoneme matching allowed to considerably mitigate this issue.
The mitigation of the recognition accuracy fall is especially important considering the use of English language that is a foreign language for the participants of the experiments, which leads easily to defective utterances or mispronounced instructions.
Table \ref{tab:accuracy} summarizes the aforementioned results for all the setups with both approaches.  

\begin{table}[t]
    \caption{Audio recognition accuracy obtained with and without phoneme matching using both Spanish and English languages.}
    \centering
    \begin{tabular}{llccc}
        \hline
        \textbf{Approach} & \textbf{Language} & \textbf{Raw input} & \textbf{Noise 5\%} & \textbf{Noise 15\%}\\
        \hline
        No phoneme  & Spanish &  97.04\% &  96.30\% & 95.56\% \\
        matching          & English &  74.81\% &  69.63\% & 59.26\% \\
                              & Both    &  85.93\% &  82.96\% & 77.41\% \\
        \hline
        With phoneme & Spanish & 100.00\% & 100.00\% & 99.26\% \\
        matching               & English &  93.33\% &  91.85\% & 78.52\% \\
                              & Both    &  96.67\% &  95.93\% & 88.89\% \\
        \hline
     \end{tabular}
     \label{tab:accuracy}
\end{table}

\begin{figure}[t]
  \centering
  \subfigure[Recognition accuracy using English instructions.] {\includegraphics[width=0.48\textwidth]{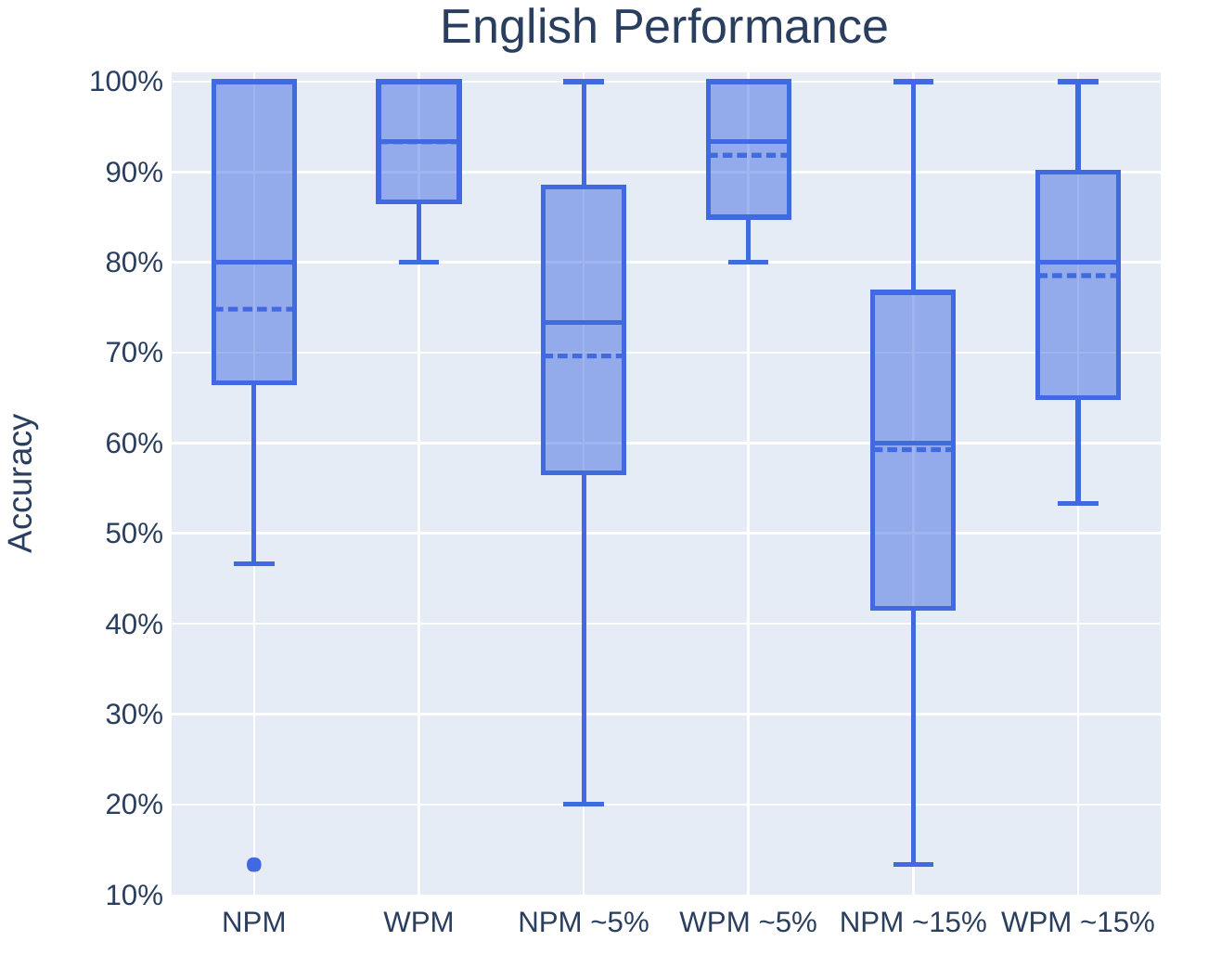} \label{fig:resultadosBoxplotENG}
  }\hfill
  \subfigure[Recognition accuracy using Spanish instructions.] {\includegraphics[width=0.48\textwidth]{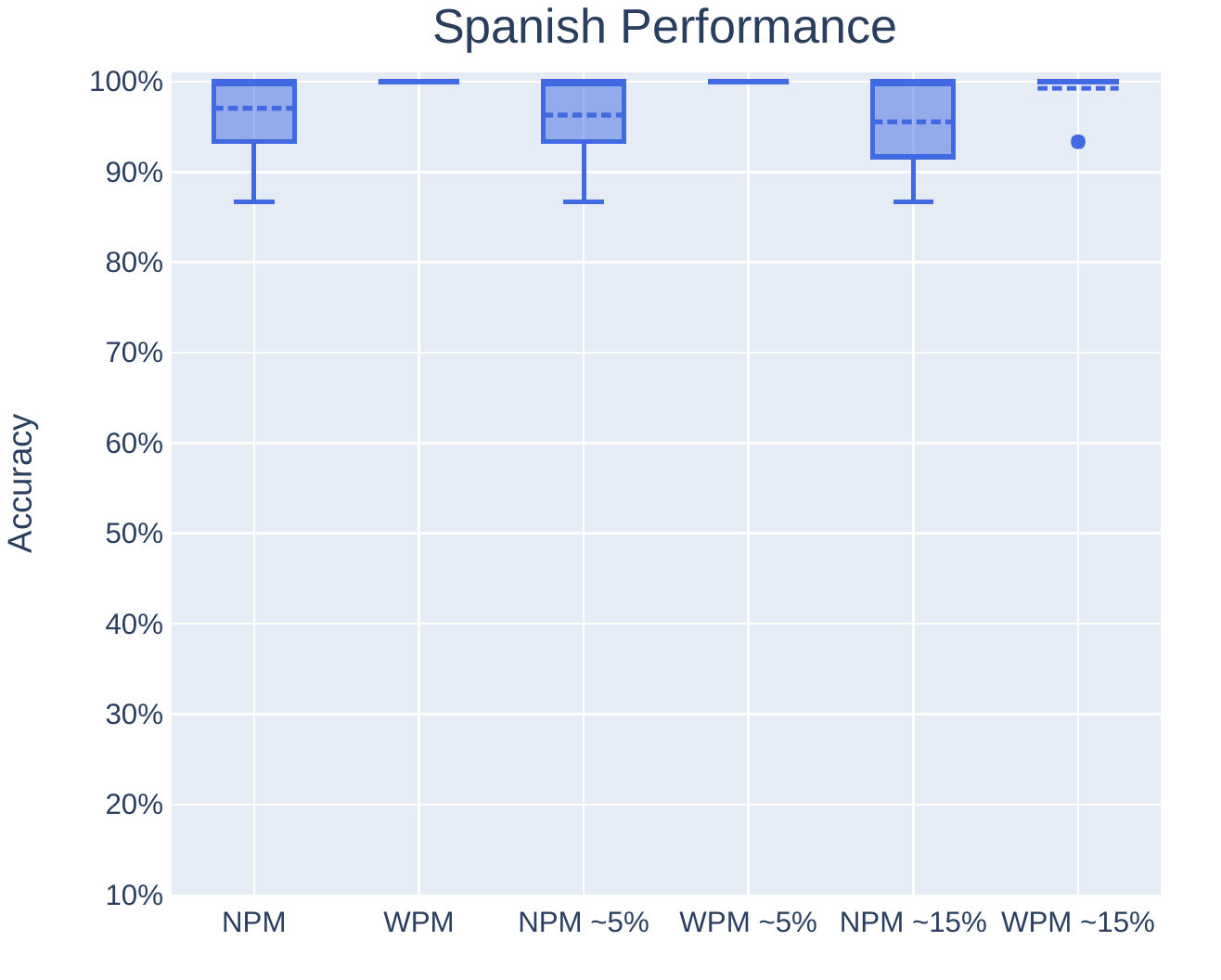} \label{fig:resultadosBoxplotSPA}
  }
  \caption{Audio recognition accuracy for all the experimental setups using both languages.
  NPM and WPM stands for no phoneme matching and with phoneme matching respectively and the percentage aside the approach represents the noise value of each setup.
  Continuous and segmented lines are used to represent the median and mean values in each box.
  The use of phoneme matching improves significantly the speech-to-action recognition even in presence of noisy inputs.} 
  \label{fig:resultadosBoxplots}
\end{figure}

Figures \ref{fig:resultadosBoxplotENG} and \ref{fig:resultadosBoxplotSPA} show the system performance as boxplots for English and Spanish instructions respectively.
The boxes are grouped considering six sets, i.e., raw inputs with no phoneme matching (NPM), raw inputs with phoneme matching (WPM), 5\% noisy inputs with no phoneme matching (NPM \string~5\%), 5\% noisy inputs with phoneme matching (WPM \string~5\%), 15\% noisy inputs with no phoneme matching (NPM \string~15\%), and 15\% noisy inputs with phoneme matching (WPM \string~15\%).
The use of English instructions leads to a larger variability among the participants of the experiments due to the participants' native language, as previously pointed out.
Although using Spanish commands overall better results are obtained, the phoneme matching technique improves the automatic speech recognition for the proposed scenario using either English or Spanish instructions.

\begin{figure}[H]
    \centering
    \includegraphics[width=\columnwidth]{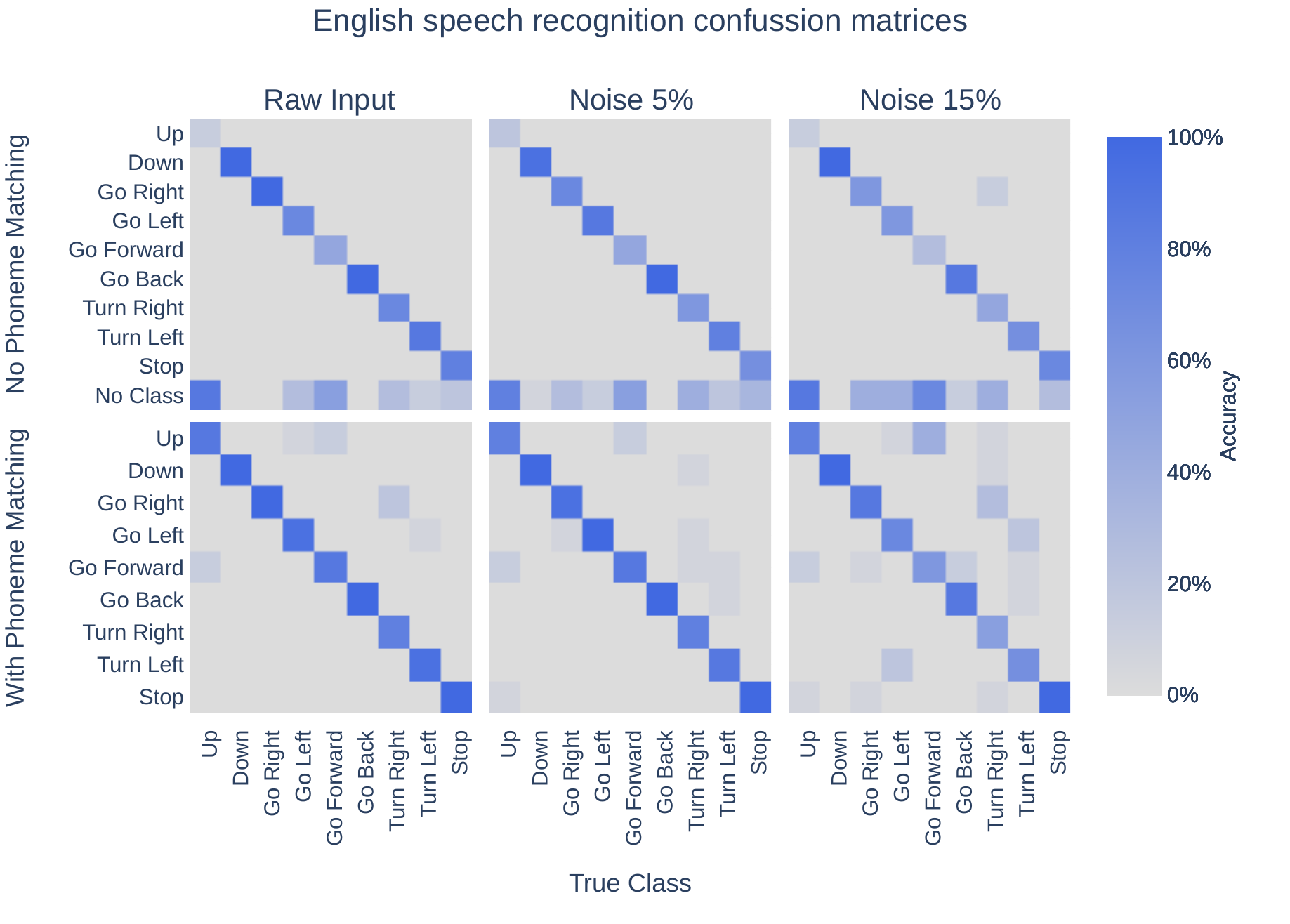}
    \caption{Predicted and true class distribution for each experimental setup using English instructions.
    The use of phoneme matching improves the overall results obtaining fewer misclassifications for all the action classes. 
    Although the use of a noisy input impoverishes the action class classification in both approaches, the use of phoneme matching allows for better recognition accuracy for all levels of noise.
    }
    \label{fig:english_matrix}
\end{figure}

\begin{figure}[t]
    \centering
    \includegraphics[width=\columnwidth]{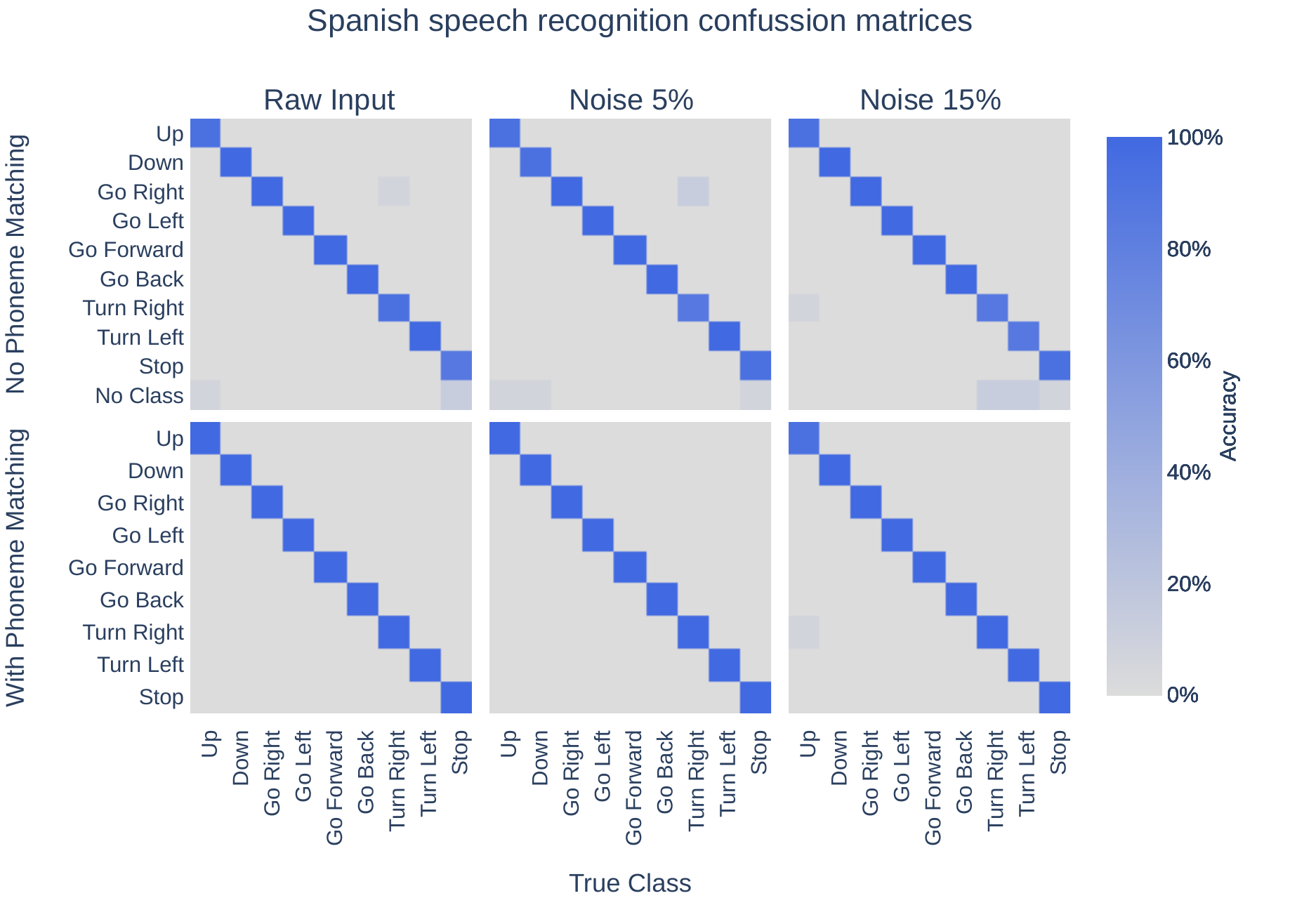}
    \caption{Predicted and true class distribution for each experimental setup using Spanish instructions.
    When Spanish language is used fewer errors in action classification are obtained in comparison to English instructions.
    Nevertheless, the use of phoneme matching still allows for better recognition accuracy for all levels of noise in comparison to the approach not using it.
    }
    \label{fig:spanish_matrix}
\end{figure}

Figure \ref{fig:english_matrix} shows the confusion matrices for the recognition of class actions using English instructions in all the experimental setups.
When no phoneme matching is used, the label “no class” refers to no coincidence between the hypothesis obtained from GCS and the instructions within the domain-based language.
Indeed, obtained results show there are many instances in which the received hypothesis does not match to any sentence in the dictionary, leading to a misclassification of the instruction. 
The implementation of phoneme matching, i.e., the algorithm computing the distance between the hypothesis received from GCS and each instruction in the domain-based dictionary, lead to a better action class recognition.
The improvement is achieved for all commands showing that the proposed approach can be used independently of the user's language ability. 
Moreover, Figure \ref{fig:spanish_matrix} shows the confusion matrices for the recognition of class actions using Spanish instructions in all the experimental setups.
In this regard, when using the user's native language, there are fewer instances of misclassification in comparison to English instructions. 
This remains similar even when a more noisy audio signal is used.

\section{Discussion}

\subsection{Summary}

In this work, we have presented an architecture to control a simulated drone through voice commands interpreted via a cloud-based automatic speech recognition system and a domain-based language.
The use of phoneme matching considerable improves the level of accuracy in instruction recognition.
Using raw inputs without phoneme matching 97.04\% and 74.81\% accuracy is obtained in action recognition for Spanish and English respectively. 
On average, voice command recognition without using phoneme matching achieves 85.93\% accuracy.
After testing the speech recognition method complemented by a domain-based language to operate the UAV in a domestic environment, better results are obtained.
The performance in instruction recognition overall improves using phoneme matching, obtaining 93.33\% and 100.00\% accuracy, for English and Spanish respectively.
On average, we obtain 96.67\% accuracy in interpreting the instructions given by the users using phoneme matching. 
Moreover, we have tested our approach in presence of 5\% and 15\% noise in the input.
Using phoneme matching, our method achieves good results in general, showing the robustness of the proposed algorithm against noise.

\subsection{Conclusions}
In conclusion, the algorithm obtains high accuracy when interpreting instructions given by an end-user through speech, being the interpretation in Spanish the one with better results. 
The main reason why Spanish interpretation results are better is that the people involved in the experiments are all Spanish native speakers.
However, the use of phoneme matching improves voice-to-action recognition, reducing the gap between languages and reaching similar results for Spanish native language users.

\subsection{Future work}
Although at this stage our approach presents some limitations such as not dealing with network interruptions or obstacle collisions, as well as being run in a simulated environment and thus keeping under control variables like noise, the obtained results motivate the extension of this work in several directions. 
For instance, a more extensive dictionary of instructions can be considered as well as the possibility of adding recognition in more languages. 
Moreover, an important next step is to transfer the proposed approach to a real-world scenario where some variables may not be easily controlled. 
In this regard, this work is the initial stage of a larger project, where we are currently developing deep reinforcement learning algorithms using interactive feedback to teach an agent on how to operate a drone.
Future extensions also take into account multi-modal sensory inputs as well as a combination of policy and reward shaping for the interactive feedback approach.

\section*{Acknowledgments}
This research was financed in part by Universidad Central de Chile under the research project CIP2018009, the Coordenação de Aperfeiçoamento de Pessoal de Nível Superior - Brasil (CAPES) - Finance Code 001, Fundação de Amparo a Ciência e Tecnologia do Estado de Pernambuco (FACEPE), and Conselho Nacional de Desenvolvimento Científico e Tecnológico (CNPq) - Brazilian research agencies.

\bibliographystyle{ieeetr}

\bibliography{references}

\end{document}